# When Ignorance is Bliss

**Peter D. Grünwald**
CWI, P.O. Box 94079
1090 GB Amsterdam
pdg@cwi.nl
http://www.grunwald.nl

**Joseph Y. Halpern**
Cornell University
Ithaca, NY 14853
halpern@cs.cornell.edu
http://www.cs.cornell.edu/home/halpern

## Abstract

It is commonly-accepted wisdom that more information is better, and that information should never be ignored. Here we argue, using both a Bayesian and a non-Bayesian analysis, that in some situations you are better off ignoring information if your uncertainty is represented by a set of probability measures. These include situations in which the information *is* relevant for the prediction task at hand. In the non-Bayesian analysis, we show how ignoring information avoids *dilation*, the phenomenon that additional pieces of information sometimes lead to an increase in uncertainty. In the Bayesian analysis, we show that for small sample sizes and certain prediction tasks, the Bayesian posterior based on a non-informative prior yields worse predictions than simply ignoring the given information.

## 1 INTRODUCTION

It is commonly-accepted wisdom that more information is better, and that information should never be ignored. Indeed, this has been formalized in a number of ways in a Bayesian framework, where uncertainty is represented by a probability measure [Good 1967; Raiffa and Shlaifer 1961]. In this paper, we argue that occasionally you are better off ignoring information if your uncertainty is represented by a set of probability measures. Related observations have been made by Seidenfeld [2004]; we compare our work to his in Section 5.

For definiteness, we focus on a relatively simple setting. Let $X$ be a random variable taking values in some set $\mathcal{X}$, and let $Y$ be a random variable taking values in some set $\mathcal{Y}$. The goal of an agent is to choose an action whose utility depends only on the value of $Y$, after having observed the value of $X$. We further assume that, before making the observation, the agent has a prior $\Pr_Y$ on the value of $Y$. If the agent actually had a prior on the joint distribution of $X$ and $Y$, then the obvious thing to do would be to condition on the observation, to get the best estimate of the value of $Y$. But we are interested in situations where the agent does not have a single prior on the joint distributions, but a family of priors $\mathcal{P}$. As the following example shows, this is a situation that arises often.

**Example 1.1:** Consider a doctor who is trying to decide if a patient has a flu or tuberculosis. The doctor then learns the patient's address. The doctor knows that the patient's address may be correlated with disease (tuberculosis may be more prevalent in some parts of the city than others), but does not know the correlation (if any) at all. In this case, the random variable $Y$ is the disease that the patient has, $\mathcal{Y} = \{\text{flu, tuberculosis}\}$, and $X$ is the neighborhood in which the agent lives. The doctor is trying to choose a treatment. The effect of the treatment depends only on the value of $Y$. Under these circumstances, many doctors would simply not take the patient's address into account, thereby ignoring relevant information. In this paper we show that this commonly-adopted strategy is often quite sensible. ∎

There is a relatively obvious sense in which ignoring information is the right thing to do. Let $\mathcal{P}$ be the set of all joint distributions on $\mathcal{X} \times \mathcal{Y}$ whose marginal on $Y$ is $\Pr_Y$. $\mathcal{P}$ represents the set of distributions compatible with the agent's knowledge. Roughly speaking, if $a^*$ is the best action given just the prior $\Pr_Y$, then then $a^*$ gives the same payoff for all joint distribution $\Pr \in \mathcal{P}$ (since they all have marginal $\Pr_Y$) We can show that every other action $a'$ will do worse than $a^*$ against *some* joint distribution $\Pr \in \mathcal{P}$. Therefore, ignoring the information leads one to adopt the minimax optimal decision. This idea is formalized as Proposition 2.1 in Section 2, where we also show that ignoring information compares very favorably to the "obvious" way of updating the set of measures $\mathcal{P}$. Proposition 2.1 makes three important assumptions:

1. There is no (second-order) distribution on the set of probabilities $\mathcal{P}$.

2. $\mathcal{P}$ contains *all* probability distributions on $\mathcal{X} \times \mathcal{Y}$ whose marginal is $\Pr_Y$.

3. The "goodness" of an action $a$ is measured by some loss or utility that—although it may be unknown to the agent at the time of updating—is *fixed*. In particular, it does not depend on the observed value of $X$.

In the remainder of the paper, we investigate the effect of dropping these assumptions. In Section 3, we consider what happens if we assume some probability distribution on the set $\mathcal{P}$ of probabilities. The obvious question is which one to use. We have to distinguish between purely subjective Bayesian approaches and so-called "noninformative", "pragmatic", or "objective" Bayesian approaches [Bernardo and Smith 1994], which are based on adopting so-called "non-informative priors". We show that for a large class of such priors, including the uniform distribution and Jeffreys' prior, using the Bayesian posterior may lead to worse decisions than using the prior $\Pr_Y$; that is, we may be better off ignoring information rather than conditioning on a noninformative prior; see Examples 3.1 and 3.2. In these examples, the posterior is based on a relatively small sample. Of course, as the sample grows larger, then using any reasonable prior will result in a posterior that converges to the true distribution. This follows directly from standard Bayesian consistency theorems [Ghosal 1998].

In Section 4 we investigate the effect of dropping the second and third assumptions. We show that once there is partial information about the relationship between $X$ and $Y$ (so that $\mathcal{P}$ is a strict subset of the set of all probability distributions on $\mathcal{X} \times \mathcal{Y}$ whose marginal is $\Pr_Y$), then the right thing to do becomes sensitive to the kind of "bookie" or "adversary" that the agent can be viewed as playing against (cf. [Halpern and Tuttle 1993]). We consider some related work, particularly that of Seidenfeld [2004], in Section 5. Our focus in this paper is on optimality in the minimax sense. It is not clear that this is the most appropriate notion of optimality. Indeed, Seidenfeld explicitly argues that it is not, and the analysis in Section 4 suggests that there are situations when ignoring information is a reasonable thing to do, even though this is not the minimax approach. We discuss alternative notions on optimality in Section 5. We conclude with further discussion in Section 6.

## 2 WHEN IGNORING HELPS: A NON-BAYESIAN ANALYSIS

In this section, we formalize our problem in a non-Bayesian setting. We then show that, in this setting, under some pragmatic assumptions, ignoring information is a sensible strategy. We also show that ignoring information compares favorably to the standard approach of working with sets of measures on $\mathcal{X} \times \mathcal{Y}$.

As we said, we are interested in an agent who must choose some action from a set $\mathcal{A}$, where the loss of the action depends only on the value of a random variable $Y$, which takes values in $\mathcal{Y}$. We assume that with each action $a \in \mathcal{A}$ and value $y \in \mathcal{Y}$ is associated some loss to the agent. (The losses can be negative, which amounts to a gain.) Let $L : \mathcal{Y} \times \mathcal{A} \to \mathbb{R} \cup \{\infty\}$ be the loss function.[1] For ease of exposition, we assume in this paper that $\mathcal{A}$ is finite.

For every action $a \in \mathcal{A}$, let $L_a$ be the random variable on $\mathcal{Y}$ such that $L_a(y) = L(y, a)$. Since $\mathcal{A}$ is assumed to be finite, for every distribution $\Pr_Y$ on $\mathcal{Y}$, there is a (not necessarily unique) action $a^* \in \mathcal{A}$ that achieves minimum expected loss, that is,

$$\inf_{a \in \mathcal{A}} (E_{\Pr_Y}[L_a]) = E_{\Pr_Y}[L_{a^*}] \qquad (1)$$

If all the agent knows is $\Pr_Y$, then it seems reasonable for the agent to choose an action $a^*$ that minimizes expected loss. We call such an action $a^*$ an *optimal* action for $\Pr_Y$.

Suppose that the agent observes the value of a variable $X$ that takes on values in $\mathcal{X}$. Further assume that, although the agent knows the marginal distribution $\Pr_Y$ of $Y$, she does not know how $Y$ depends on $X$. That is, the agent's uncertainty is characterized by the set $\mathcal{P}$ consisting of all distributions on $\mathcal{X} \times \mathcal{Y}$ with marginal distribution $\Pr_Y$ on $\mathcal{Y}$. The agent now must choose a *decision rule* that determines what she does as a function of her observations. We allow decision rules to be randomized. Thus, if $\Delta(\mathcal{A})$ consists of all probability distributions on $\mathcal{A}$, a decision rule is a function $\delta : \mathcal{X} \to \Delta(\mathcal{A})$ that chooses a distribution over actions based on her observations. Let $\mathcal{D}(\mathcal{X}, \mathcal{A})$ be the set of all such decision rules. A special case is a deterministic decision rule, that assigns probability 1 to a particular action. If $\delta$ is deterministic, we sometimes abuse notation and write $\delta(x)$ for the action that is assigned probability 1 by the distribution $\delta(x)$. Given a decision rule $\delta$ and a loss function $L$, let $L_\delta$ be the random variable on $\mathcal{X} \times \mathcal{Y}$ such that $L_\delta(x, y) = \sum_{a \in \mathcal{A}} \delta(x)(a) L(y, a)$. Here $\delta(x)(a)$ stands for the probability of performing action $a$ according to the distribution $\delta(x)$ over actions that is adopted when $x$ is observed. Note that in the special case that $\delta$ is a deterministic decision rule, then $L_\delta(x, y) = L(y, \delta(x))$. Moreover, if $\delta_a$ is the (deterministic) decision rule that always chooses $a$, then $L_{\delta_a}(x, y) = L_a(y)$.

The following result, whose proof we leave to the full paper, shows that the decision rule $\delta^*$ that always chooses an optimal action $a^*$ for $\Pr_Y$, independent of the observation, is optimal in a minimax sense. Note that the worst-case expected loss of decision-rule $\delta$ is $\sup_{\Pr \in \mathcal{P}} E_{\Pr}[L_\delta]$. Thus, the best worst-case loss (i.e., the *minimax* loss) over all decision rules is $\inf_{\delta \in \mathcal{D}(\mathcal{X}, \mathcal{A})} \sup_{\Pr \in \mathcal{P}} E_{\Pr}[L_\delta]$.

---

[1]We could equally well use utilities, which can be viewed as a positive measure of gain. Losses seem to be somewhat more standard in this literature.

**Proposition 2.1:** *Suppose that* $\Pr_Y$ *is an arbitrary distribution on* $\mathcal{Y}$, $L$ *is an arbitrary loss function,* $\mathcal{P}$ *consists of all distributions on* $\mathcal{X} \times \mathcal{Y}$ *with marginal* $\Pr_Y$, *and* $a^*$ *is an optimal action for* $\Pr_Y$ *(with respect to the loss function* $L$*). Then* $E_{\Pr_Y}[L_{a^*}] = \inf_{\delta \in \mathcal{D}(\mathcal{X},\mathcal{A})} \sup_{\Pr \in \mathcal{P}} E_{\Pr}[L_\delta]$.

A standard decision rule when uncertainty is represented by a set $\mathcal{P}$ of probability measures is the Maxmin Expected Utility Rule [Gilboa and Schmeidler 1989]; compute the expected utility (or expected loss) of an action with respect to each of the probability measures in $\mathcal{P}$, and then choose the action whose worst-case expected utility is best (or worst-case expected loss is least). Proposition 2.1 says that if $\mathcal{P}$ consists of all probability measures with marginal $\Pr_Y$ and the loss depends only on the value of $Y$, then the action with the least worst-case loss is an optimal action with respect to $\Pr_Y$.

**Example 2.2:** Consider perhaps the simplest case, where $\mathcal{X} = \mathcal{Y} = \{0, 1\}$. Suppose that our agent knows that $E_{\Pr_Y}[Y] = \Pr_Y(Y = 1) = p$ for some fixed $p$. As before, let $\mathcal{P}$ be the set of distributions on $\mathcal{X} \times \mathcal{Y}$ with marginal $\Pr_Y$. Suppose further that the only actions are 0 and 1 (intuitively, these actions amount to predicting the value of $Y$), and that the loss function is 0 if the right value is predicted and 1 otherwise; that is, $L(i,j) = |i-j|$. This is the so-called 0/1 or *classification* loss. It is easy to see that $E[L_0] = p$ and $E[L_1] = 1-p$, so the optimal act is to choose 0 if $p < .5$ and 1 if $p > .5$ (both acts have loss $1/2$ if $p = .5$). The loss of the optimal act is $\min(p, 1-p)$.

Perhaps the more standard approach for dealing with uncertainty in this case is to work with the whole set of distributions. Assume that $0 < \Pr(Y = 1) = p < 1$. Let $\mathcal{P}_i = \{\Pr(\cdot \mid X = i) : \Pr \in \mathcal{P}\}$, $i = 0,1$. Then for all $q \in [0,1]$, both $\mathcal{P}_0$ and $\mathcal{P}_1$ contain a distribution $\Pr_q$ such that $\Pr_q(Y = 1) = q$. In other words, $\mathcal{P}_0 = \mathcal{P}_1 = \Delta(\mathcal{Y})$, the set of all distributions on $\mathcal{Y}$. Observing $X = x$ causes all information about $Y$ to be lost. Remarkably, this holds *no matter what value of $X$ is observed*.

Thus, even though the agent knew that $\Pr(Y = 1) = p$ before observing $X$, after observing $X$, the agent has no idea of the probability that $Y = 1$. This is a special case of a phenomenon that has been called *dilation* in the statistical and imprecise probability literature [Augustin 2003; Cozman and Walley 2001; Herron, Seidenfeld, and Wasserman 1997; Seidenfeld and Wasserman 1993]: it is possible that lower probabilities strictly decrease and that upper probabilities strictly increase, no matter what value of $x$ is observed. Dilation has severe consequences for decision-making. The minimax-optimal decision rule $\delta^*$ with respect to $\mathcal{P}_Y$ is to randomize, choosing both 0 and 1 with probability $1/2$. Note that, no matter what $\Pr \in \mathcal{P}$ actually obtains,

$$E_{\Pr}[L_{a^*}] = \min\{p, 1-p\} \ ; \ E_{\Pr}[L_{\delta^*}] = 1/2.$$

Thus, if $p$ is close to 0 or 1, ignoring information does much better than making use of it.

This can be viewed as an example of what decision theorists have called *time inconsistency*. Suppose, for definiteness, that $p = 1/3$. Then, a priori, the optimal strategy is to decide 0 no matter what. On the other hand, if either $X = 0$ or $X = 1$ is observed, then the optimal action is to randomize. When uncertainty is described with a single probability distribution (and updating is done by conditioning), then time inconsistency cannot occur.[2] ∎

## 3 WHEN IGNORING HELPS: A BAYESIAN ANALYSIS

Suppose that, instead of having just a set $\mathcal{P}$ of probability measures, the agent has a probability measure on $\mathcal{P}$. But then which probability measure should she take? Broadly speaking, there are two possibilities here. We can consider either *purely subjective* Bayesian agents or *pragmatic* Bayesian agents. A purely subjective Bayesian agent will come up with some (arbitrary) prior that expresses her subjective beliefs about the situation. It then makes sense to assess the consequences of ignoring information in terms of expected loss, where the expectation is taken with respect to the agent's subjective prior. Good's total evidence theorem, a classical result of Bayesian decision theory [Good 1967; Raiffa and Shlaifer 1961], states that, when taking the expectation with respect to the agent's prior, the optimal decision should always be based on conditioning on *all* the available information—information should never be ignored.

In contrast, we consider an agent who adopts Bayesian updating for pragmatic reasons (i.e., because it usually works well) rather than for fundamental reasons. In this case, because computation time is limited and/or prior knowledge is hard to obtain or formulate, the prior adopted is typically easily computable and "noninformative", such as a prior that is uniform in some natural parameterization of $\mathcal{P}$. We suspect that many statisticians are pragmatic Bayesians in this sense. (Indeed, most "Bayesian" UAI and statistics papers adopt pragmatic priors that cannot seriously be viewed as fully subjective.) When analyzing such a pragmatic approach, it no longer makes that much sense to compare ignoring information to Bayesian updating on new information by looking at the expected loss with respect to the adopted prior. The reason is that the prior can no longer be expected to correctly reflect the agent's degrees of belief. It seems more meaningful to pick a single probability measure $\Pr$ and to analyze the behavior of the Bayesian under the assumption that $\Pr$ is the "true" state of nature. By varying $\Pr$ over the set $\mathcal{P}$, we can get a sense of the behav-ior of Bayesian updating in all possible situations. This is the type of analysis that we adopt in this section; it is quite standard in the statistical literature on consistency of Bayes methods [Blackwell and Dubins 1962; Ghosal 1998].

---

[2]It has been claimed that this time consistency also depends on the agent having perfect recall; see [Halpern 1997; Piccione and Rubinstein 1997] for some discussion of this issue.

We focus on a large class of priors on $\mathcal{P}$ that includes most standard recommendations for noninformative priors. Essentially, we show that for any prior in the class, when the sample size is small, ignoring information is better than using the Bayesian posterior. That is, if a pragmatic agent has the choice between (a) first adopting a pragmatic prior, perhaps not correctly reflecting her own beliefs, and then reasoning like a Bayesian, or, (b) simply ignoring the available information, then, when the sample size is small, she might prefer option (b). On the other hand, as more information becomes available, the Bayesian posterior behaves almost as well as ignoring the information in the worst case, and substantially better than ignoring in most other cases. (Of course, part of the issue here is what counts as "better" when uncertainty is represented by a set of probability measures. For the time being, we say that "A is better than B" if A achieves better minimax behaviour than B. We return to this issue at the end of this section.)

**Example 3.1:** As in Example 2.2, let $\mathcal{X} = \mathcal{Y} = \{0, 1\}$. For definiteness, suppose that the known prior $\Pr_Y$ is such that $\Pr_Y(Y = 1) = p$. Throughout this section we assume that $0 < p < 1$. A probability measure on $\mathcal{X} \times \mathcal{Y}$ is completely determined by $\Pr(X = 1 \mid Y = 1)$ and $\Pr(X = 1 \mid Y = 0)$. Moreover, for every choice $(\alpha, \beta) \in [0, 1] \times [0, 1]$ for these two conditional probabilities, there is a probability $\Pr_{\alpha,\beta} \in \mathcal{P}$; in fact

$$\Pr_{\alpha,\beta}(X = 1, Y = 1) = p\alpha;$$
$$\Pr_{\alpha,\beta}(X = 1, Y = 0) = (1-p)\beta;$$
$$\Pr_{\alpha,\beta}(X = 0, Y = 1) = p(1-\alpha);$$
$$\Pr_{\alpha,\beta}(X = 0, Y = 0) = (1-p)(1-\beta).$$

Notice that $\Pr_{\alpha,\beta}(X = 1) = p\alpha + (1-p)\beta$. Given this, one obvious way to put a uniform prior on $\mathcal{P}$ is just to take a uniform prior on the square $[0,1]^2$; we adopt this prior for the time being and consider other notions of "uniform" further below.

To calculate the Bayesian predictions of $Y$ given $X$, we must first determine the Bayesian "marginal" probability measure $\overline{\Pr}$, where $\overline{\Pr}(X = i, Y = j) = \int_{\alpha=0}^1 \int_{\beta=0}^1 \Pr_{\alpha,\beta}(X = i, Y = j) d\alpha d\beta$ ("marginal" because we are marginalizing out the parameters $\alpha$ and $\beta$), and then use $\overline{\Pr}$ to calculate the expected loss of predicting $Y = 1$. That is, we calculate the so-called "predictive distribution" $\overline{\Pr}(Y = \cdot \mid X = \cdot)$. We can calculate this directly without performing any integration as follows. By symmetry, we must have $\overline{\Pr}(Y = 1 \mid X = 1) = \overline{\Pr}(Y = 1 \mid X = 0)$. Now if $\gamma = \overline{\Pr}(X = 1)$, then it must be the case that $\gamma \overline{\Pr}(Y = 1 \mid X = 1) + (1 - \gamma)\overline{\Pr}(Y = 1 \mid X = 0) = p$; this implies that $\overline{\Pr}(Y = 1 \mid X = 1) = p$. Thus, when calculating the predictive distribution of $Y$ after observing $X$, the Bayesian will always ignore the value of $X$ and predict with his marginal distribution $\overline{\Pr}_Y$.

Thus, before observing data, the Bayesian ignores the value of $X$, and thus makes minimax-optimal decisions. Potentially suboptimal behavior of the Bayesian can occur only *after* the Bayesian has observed some data. To analyze this case, we need to assume that we have a sequence of $n$ observations $(X_1, Y_1), \ldots, (X_n, Y_n)$ and are trying to predict the value of $Y_{n+1}$, given the value of $X_{n+1}$. The distribution $\Pr_{\alpha,\beta}$ on $\mathcal{X} \times \mathcal{Y}$ is extended to a distribution $\Pr^n_{\alpha,\beta}$ on $(\mathcal{X} \times \mathcal{Y})^n$ by assuming that the observations are independent. Of course, the hope is that the observations will help us learn about $\alpha$ and $\beta$, allowing us to make better decisions on $Y$. To take the simplest case, suppose that we have observed that $(X_1, Y_1) = (1, 1)$, and $X_2 = 1$, and want to calculate the value of $Y_2$. Note that

$$\begin{aligned}
&\overline{\Pr}^2(Y_2 = 1 \mid X_2 = 1, (X_1, Y_1) = (1,1)) \\
&= \frac{\overline{\Pr}^2((X_2,Y_2)=(1,1),(X_1,Y_1)=(1,1))}{\sum_{y\in\{0,1\}} \overline{\Pr}^2((X_2,Y_2)=(1,y),(X_1,Y_1)=(1,1))} \\
&= \frac{\int_{\alpha=0}^1 \int_{\beta=0}^1 (p\alpha)^2 d\alpha d\beta}{\int_{\alpha=0}^1\int_{\beta=0}^1 (p\alpha)^2 d\alpha d\beta + \int_{\alpha=0}^1\int_{\beta=0}^1 (p\alpha)(1-p)\beta d\alpha d\beta} \\
&= \frac{\frac{1}{3}p}{\frac{1}{3}p + (1-p)\frac{1}{4}} \\
&= \frac{4p}{p+3}.
\end{aligned}$$

Since we must have $\overline{\Pr}^2(Y_2 = 1 \mid (X_1, Y_1) = (1, 1)) = p$, it follows using the same symmetry argument as above that if $(X_1, Y_1) = (1, 1)$, then, no matter what value of $X_2$ is observed, the value of $X_2$ is not ignored. Similar calculations show that, if $(X_1, Y_1) = (i, j)$ for all $i, j \in \{0, 1\}$, then, no matter what value of $X_2$ is observed, the value of $X_2$ is not ignored. ∎

In Example 3.1 we claimed that the Bayesian should predict by the predictive distribution $\overline{\Pr}^2(Y = 1 \mid X = 1)$ as defined in the example. While this is the standard Bayesian approach, one may also directly consider the "expected" conditional probability $\int_{\alpha=0}^1 \int_{\beta=0}^1 \Pr^2_{\alpha,\beta}(Y = 1 \mid X = 1)$. These two approaches give different answers, since expectation does not commute with division. To see why we prefer the standard approach, note that, because of independence, $\Pr^2_{\alpha,\beta}(Y_2 = i \mid X_2 = j, (X_1, Y_1) = (i', j')) = \Pr_{\alpha,\beta}((X_1, Y_1) = (i', j')) \Pr_{\alpha,\beta}(Y_2 = i \mid X_2 = j)$; similarly with repeated observations. That is, with the alternative approach, there would be no learning from data. Thus, for the remainder of the paper, we use the predictive-distribution approach, with no further comment.

**Example 3.2:** Now consider the more general situation where $\mathcal{X} = \{1, \ldots, M\}$ for arbitrary $M$, and $\mathcal{Y} = \{0, 1\}$ as before. We consider a straightforward extension of the previous set of distributions: let $\vec{\alpha} = (\alpha_1, \ldots, \alpha_M)$ be an element of the $M$-dimensional unit simplex; $\vec{\beta}$ is defined similarly. Fix $p \in [0, 1]$, and define

$$\Pr_{\vec{\alpha}, \vec{\beta}}(Y = 1) = p;\ \Pr_{\vec{\alpha}, \vec{\beta}}(X = j \mid Y = 1) = \alpha_j;$$
$$\Pr_{\vec{\alpha}, \vec{\beta}}(X = j \mid Y = 0) = \beta_j.$$

Note that
$$\Pr_{\vec{\alpha}, \vec{\beta}}(X = j, Y = 1) = \alpha_j p$$

and
$$\Pr_{\vec{\alpha}, \vec{\beta}}(X = j, Y = 0) = \beta_j (1 - p).$$

Let $D$ be a random variable used to denote the outcome of then $n$ observations $(X_1, Y_1), \ldots, (X_n, Y_n)$. Given a sequence $(\vec{x}, \vec{y}) = ((x_1, y_1), \ldots, (x_n, y_n))$ of observations, let $n_k^{(\vec{x}, \vec{y})}$ denote the number of observations in the sequence with $Y_i = k$, for $k \in \{0, 1\}$. Similarly, $n_{(j,k)}^{(\vec{x}, \vec{y})}$ denotes the number of observations $(X_i, Y_i)$ in the sequence with $(X_i = j, Y_i = k)$. Then

$$\Pr_{\vec{\alpha}, \vec{\beta}}^n(D = (\vec{x}, \vec{y}))$$
$$= p^{n_1^{(\vec{x}, \vec{y})}} (1-p)^{n_0^{(\vec{x}, \vec{y})}} \prod_{j=1}^M \alpha_j^{n_{(j,1)}^{(\vec{x}, \vec{y})}} \prod_{j=1}^M \beta_j^{n_{(j,0)}^{(\vec{x}, \vec{y})}}.$$

We next put a prior on $\mathcal{P} = \{\Pr_{\vec{\alpha}, \vec{\beta}} : \alpha, \beta \in [0, 1]\}$ We restrict attention to priors that can be written as a product of *Dirichlet distributions* [Bernardo and Smith 1994]. A Dirichlet distribution on the $M$-dimensional unit simplex $\Delta_M$ (which we can identify with the set of probability distributions on $\{1, \ldots, M\}$) is parameterized by an $M$-dimensional vector $\vec{a}$. For $\vec{a} = (a_1, \ldots, a_M)$, the $\vec{a}$-Dirichlet distribution has density $p_{\vec{a}}$ that satisfies, for all $\vec{\alpha} \in \Delta_M$,

$$p_{\vec{a}}(\vec{\alpha}) = \frac{1}{Z(\vec{a})} \alpha_1^{a_1 - 1} \cdot \ldots \cdot \alpha_M^{a_M - 1},$$

where $Z(\vec{a}) = \int_{\vec{\alpha} \in \Delta_M} \alpha_1^{a_1 - 1} \cdot \ldots \cdot \alpha_M^{a_M - 1} d\vec{\alpha}$ is a normalizing factor. Note that the uniform prior is the $\vec{a}$-Dirichlet prior where $a_1 = a_2 = \ldots = a_M = 1$. As we shall see, many other priors of interest are special cases of Dirichlet priors.

We consider only priors $w$ on $\mathcal{P}$ that satisfy $w(\vec{\alpha}, \vec{\beta}) = w_{\vec{a}}(\vec{\alpha}) w_{\vec{b}}(\vec{\beta})$ for all $\vec{\alpha}, \vec{\beta} \in \Delta_M$, where $w_{\vec{a}}$ and $w_{\vec{b}}$ are of the Dirichlet form. Then

$$\overline{\Pr}^n(D = (\vec{x}, \vec{y}))$$
$$= \int_{\vec{\alpha} \in \Delta_M} \int_{\vec{\beta} \in \Delta_M} \Pr_{\vec{\alpha}, \vec{\beta}}^n(D = (\vec{x}, \vec{y})) w_{\vec{a}}(\vec{\alpha}) w_{\vec{b}}(\vec{\beta}) d\vec{\alpha} d\vec{\beta}. \quad (2)$$

Now suppose that a Bayesian has observed an initial sample $D$ of size $n$ and $X_{n+1}$, and must predict $Y_{n+1}$. Suppose $X_{n+1} = k$. Then the Bayesian's predictive distribution becomes $\overline{\Pr}^{n+1}(Y_{n+1} = \cdot \mid X_{n+1}, D)$ or, more explicitly,

$$\overline{\Pr}^{n+1}(Y_{n+1} = j \mid X_{n+1} = k, D = (\vec{x}, \vec{y}))$$
$$= \frac{\overline{\Pr}^{n+1}(D = (\vec{x}, \vec{y}), X_{n+1} = k, Y_{n+1} = j)}{\overline{\Pr}^{n+1}(D = (\vec{x}, \vec{y}), X_{n+1} = k)}.$$

It will be convenient to represent this distribution by the odds ratio

$$\frac{\overline{\Pr}^{n+1}(Y_{n+1} = 1 \mid X_{n+1} = k, D = (\vec{x}, \vec{y}))}{\overline{\Pr}^{n+1}(Y_{n+1} = 0 \mid X_{n+1} = k, D = (\vec{x}, \vec{y}))}$$
$$= \frac{\overline{\Pr}^{n+1}(D = (\vec{x}, \vec{y}), X_{n+1} = k, Y_{n+1} = 1)}{\overline{\Pr}^{n+1}(D = (\vec{x}, \vec{y}), X_{n+1} = k, Y_{n+1} = 0)}. \quad (3)$$

Both the numerator and the denominator of the right-hand side of (3) are of the form (2), so this expression is a ratio of Dirichlet integrals. These can be calculated explicitly [Bernardo and Smith 1994], giving

$$\frac{\overline{\Pr}^{n+1}(Y_{n+1} = 1 \mid X_{n+1} = k, D = (\vec{x}, \vec{y}))}{\overline{\Pr}^{n+1}(Y_{n+1} = 0 \mid X_{n+1} = k, D = (\vec{x}, \vec{y}))}$$
$$= \frac{\overline{\Pr}^{n+1}(D = (\vec{x}, \vec{y}), X_{n+1} = k, Y_{n+1} = 1)}{\overline{\Pr}^{n+1}(D = (\vec{x}, \vec{y}), X_{n+1} = k, Y_{n+1} = 0)} \quad (4)$$
$$= \frac{p}{1-p} \cdot \frac{n_{(k,1)}^{(\vec{x}, \vec{y})} + a_k}{n_{(k,0)}^{(\vec{x}, \vec{y})} + b_k} \cdot \frac{n_0^{(\vec{x}, \vec{y})} + \sum_{k=1}^M b_k}{n_1^{(\vec{x}, \vec{y})} + \sum_{k=1}^M a_k}.$$

With the uniform prior, (4) simplifies to

$$\frac{p}{1-p} \cdot \frac{n_{(k,1)}^{(\vec{x}, \vec{y})} + 1}{n_{(k,0)}^{(\vec{x}, \vec{y})} + 1} \cdot \frac{n_0^{(\vec{x}, \vec{y})} + M}{n_1^{(\vec{x}, \vec{y})} + M}. \quad (5)$$

(4) and (5) show that the odds-ratio behaves like $p/(1-p)$ (which would be the odds-ratio obtained by ignoring the values of $X$) times some "correction factor". Ideally this correction factor would be close to 1 for small samples and then smoothly change "in the right direction", so that the Bayesian's predictions are never much worse than the minimax predictions and, as more data comes in, get monotonically better and better. We now consider two examples to show the extent to which this happens.

First, take $M = 2$, and let Pr be such that $\Pr(Y = 1) = p$, $\Pr(X = 1 \mid Y = 1) = 1$, and $\Pr(X = 0 \mid Y = 0) = 1$. Then, for $k = 1$, (4) becomes

$$\frac{\overline{\Pr}^{n+1}(Y_{n+1} = 1 \mid X_{n+1} = 1, D = (\vec{x}, \vec{y}))}{\overline{\Pr}^{n+1}(Y_{n+1} = 0 \mid X_{n+1} = 1, D = (\vec{x}, \vec{y}))}$$
$$= \frac{p}{1-p} \cdot \frac{n_1^{(\vec{x}, \vec{y})} + 1}{1} \cdot \frac{n_0^{(\vec{x}, \vec{y})} + 2}{n_1^{(\vec{x}, \vec{y})} + 2}.$$

For all but the smallest $n$, with high Pr-probability, $n_1^{(\vec{x}, \vec{y})} \approx pn$. Thus, the odds ratio tends to infinity, as expected.

In the previous example, $X$ and $Y$ were completely correlated. Suppose that they are independent. That is, suppose again that $M = 2$, but that Pr is such that $\Pr(Y = 1 \mid X = k) = p$, for $k = 0, 1$. For simplicity, we further suppose that $p = 1/2$ and that $\Pr(X = 0) = \Pr(X = 1) = 1/2$; the same argument applies with little change if we drop these assumptions.

Given $\alpha > 1$, consider a loss function $L_\alpha$ with asymmetric misclassification costs, given by $L_\alpha(0, 0) = L_\alpha(1, 1) = 0$; $L_\alpha(1, 0) = 1$; $L_\alpha(0, 1) = \alpha$. Clearly, $E_{\Pr_Y}[L_0] = 0.5$ and $E_{\Pr_Y}[L_1] = 0.5\alpha$. Thus, the optimal action with respect to the prior $\Pr_Y$ is to predict 0, and the minimax-optimal action is to always predict 0. Moreover, the expected loss of predicting 1 is $.5(\alpha - 1)$.

Now consider the predictions of a Bayesian who uses the uniform prior. The Bayesian will predict 1 iff

$$\frac{E_{\overline{\Pr}(Y_{n+1}|X_{n+1},D)}[L_1]}{E_{\overline{\Pr}(Y_{n+1}|X_{n+1},D)}[L_0]}$$
$$= \frac{\alpha \overline{\Pr}(Y_{n+1}=0|X_{n+1}=k,D=(\vec{x},\vec{y}))}{\overline{\Pr}(Y_{n+1}=1|X_{n+1}=k,D=(\vec{x},\vec{y}))} < 1.$$

From the odds-ratio (5) we see that this holds iff

$$\begin{aligned}\alpha &< \frac{\overline{\Pr}(Y_{n+1}=1|X_{n+1}=k,D=(\vec{x},\vec{y}))}{\overline{\Pr}(Y_{n+1}=0|X_{n+1}=k,D=(\vec{x},\vec{y}))} \\ &= \frac{n^{(\vec{x},\vec{y})}_{(k,1)}+1}{n^{(\vec{x},\vec{y})}_{(k,0)}+1} \cdot \frac{n^{(\vec{x},\vec{y})}_0+2}{n^{(\vec{x},\vec{y})}_1+2}.\end{aligned} \quad (6)$$

If $\beta$ is the probability (with respect to $\Pr^n$) of (6), then the difference between the Bayesian's expected loss and the expected loss of someone who ignores the information is $\beta(\alpha - 1)/2$. Clearly, $\beta$ depends on $\alpha$ and $n$. Moreover, for any fixed $\alpha > 1$, $\lim_{n \to \infty} \beta \to 0$. This, of course, just says that eventually the Bayesian will learn correctly. However, for relatively small $n$, it is not hard to construct situations where $\beta(\alpha-1)/2$ can be nontrivial. For example, if $n = 4$ and $\alpha = 1.4$, then $\beta \sim .35$. (We computed this by a brute force calculation, by considering all the values $n^{(\vec{x},\vec{y})}_{(k,j)}$ that cause (6) to be true, and computing their probability.) Thus, the Bayesian's expected loss is about 14% worse than that of an agent who ignores the information.

Although in this example there is no dependence between $X$ and $Y$ in the actual distribution, by continuity, the same result holds if there is some dependence. ∎

This conclusion assumed that a Bayesian chose a particular noninformative prior, but it does not depend strongly on this choice. As is well known, there is no unique way of defining a "uniform prior" on a set of distributions $\mathcal{P}$, since what is "uniform" depends on the chosen parameterization. For this reason, people have developed other types of noninformative priors. One of the most well-known of these is the so-called Jeffreys' prior [Jeffreys 1946; Bernardo and Smith 1994], specifically designed as a prior expressing "ignorance". This prior is invariant under continuous 1-to-1 reparameterizations of $\mathcal{P}$. It turns out that Jeffreys' prior on the set $\mathcal{P}$ is also of the Dirichlet form (with $a_1 = \ldots = a_n = b_1 = \ldots = b_n = 1/2$) so that it satisfies (4) (see, for example, [Kontkanen, Myllymäki, Silander, Tirri, and Grünwald 2000]). Other pragmatic priors that are often used in practice are the so-called *equivalent sample size (ESS) priors* [Kontkanen, Myllymäki, Silander, Tirri, and Grünwald 2000]. For the case of our $\mathcal{P}$, these also take the Dirichlet form. Thus, the analysis of Example 3.2 does not substantially change if we use the Jeffreys' prior or an ESS prior. It remains the case that, for certain sample sizes, ignoring information is preferable to using the Bayesian posterior.

Example 3.2 shows that with noninformative priors, for small sample sizes, ignoring the information may be better than Bayesian updating. Essentially, the reason for this is that all standard noninformative priors assign probability zero to the set of distributions $\mathcal{P}' \subseteq \mathcal{P}$ according to which $X$ and $Y$ are independent. But the measures in $\mathcal{P}'$ are exactly the ones that lead to minimax-optimal decisions. Of course, there is no reason that a Bayesian must use a noninformative prior. In some settings it may be preferable to adopt a "hierarchical pragmatic prior" that puts a uniform probability on both $\mathcal{P} - \mathcal{P}'$ and $\mathcal{P}'$, and assigns probability 0.5 to each of $\mathcal{P} - c\mathcal{P}'$ and $\mathcal{P}'$. Such a prior makes it easier for a Bayesian to learn that $X$ and $Y$ are independent. (A closely related prior has been used by Barron, Rissanen, and Yu [1998], in the context of universal coding, with a logarithmic loss function.) With such a prior, a Bayesian would do better in this example.

The notion of optimality that we have used up to now is minimax loss optimality. Prediction $i$ is better than prediction $j$ if the worst-case loss when predicting $i$ (taken over all possible priors in $\mathcal{P}$) is better than the worst-case loss when predicting $j$. But there are certainly other quite reasonable criteria that could be used when comparing predictions. In particular, we could consider minimax regret. That is, we could consider the prediction that minimizes the worst-case difference between the best prediction for each $\Pr \in \mathcal{P}$ and the actual prediction. In the second half of Example 3.2, we calculated that if the true probability $\Pr$ is such that $\Pr(X = 0) = \Pr(X = 1) = 1/2$ and $\Pr$ makes $X$ and $Y$ independent, then the difference between the loss incurred by an agent that ignores the prior and a Bayesian is roughly .07. We do not know if there are probabilities $\Pr'$ for which the Bayesian agent does much worse than an agent who ignores the prior with respect to $\Pr'$. On the other hand, if $X$ and $Y$ are completely correlated, that is, if the true probability $\Pr$ is such that $\Pr(Y = 1 \mid X = 1) = \Pr(Y = 0 \mid X = 0) = 1$, then if $n = 4$ and $\alpha = 1.4$, the Bayesian will predict correctly, while half the time the agent that ignores information will not. Then the difference between the loss incurred by the Bayesian agent and the agent that ignores the information is 0.5. Thus, in the sense of expected regret, the Bayesian approach is bound to be at least as good as ignoring the information in this example. We are currently investigating whether this is true more generally.

## 4 PARTIAL IGNORANCE AND DIFFERENT TYPES OF BOOKIES

In Sections 2 and 3, we showed that ignoring information is sensible as long as (1) the set $\mathcal{P}$ contains *all* distributions on $\mathcal{X} \times \mathcal{Y}$ with the given marginal $\Pr_Y$, and (2) the loss function $L$ is fixed; in particular, it does not depend on the realized value of $X$. In this section, we consider what happens when we drop these assumptions.

The assumption that $\mathcal{P}$ contains all distributions on $\mathcal{X} \times \mathcal{Y}$ with the given marginal $\Pr_Y$ amounts to the assumption that all the agent knows is $\Pr_Y$. If an agent has more information about the probability distribution on $\mathcal{X} \times \mathcal{Y}$, then ignoring information is in general not a reasonable thing to do. To take a simple example, suppose that the set $\mathcal{P}$ contains only one distribution $P^\circ$. Then clearly the minimax optimal strategy is to use the decision rule based on the conditional distribution $P^\circ(Y \mid X)$, which means that all available information is taken into account. Using $P^\circ_Y$ is clearly not the right thing to do.

On the other hand, if $\mathcal{P}$ is neither a singleton nor the set of all distributions with the given marginal $\Pr_Y$, then ignoring may or may not be minimax optimal, depending on the details of $\mathcal{P}$. Even in some cases where ignoring is not minimax optimal, it may still be a reasonable update rule to use, because, no matter what $\mathcal{P}$ is, ignoring $X$ is a *reliable* update rule [Grünwald 2000]. This means the following: Suppose that the loss function $L$ is known to the agent. Let $a^*$ be the optimal action resulting from ignoring information about $X$, that is, adopting the marginal $\Pr_Y$ as the distribution of $Y$, independently of what $X$ was observed. Then it must be the case that

$$E_{\Pr_Y}[L_{a^*}] = E_{\Pr}^{(X,Y)}[L_{a^*}(X,Y)], \qquad (7)$$

meaning that the loss the *agent* expects to have using his adopted action $a^*$ is guaranteed to be identical to the *true* expected loss of the agent's action $a^*$. Thus, the quality of the agent's predictions is exactly as good as the agent thinks they are, and the agent cannot be overly optimistic about his own performance. Data will behave as if the agent's adopted distribution $\Pr_Y$ is correct, even though it is not.

This desirable property of reliability is lost when the loss function can depend on the observation $X$. To understand the impact of this possibility, consider again the situation of Example 2.2, except now assume that the loss function can depend on the observation.

**Example 4.1:** As in Example 2.2, assume that $\mathcal{X} = \mathcal{Y} = \{0, 1\}$, that the agent knows that $E_{\Pr_Y}[Y] = \Pr_Y(Y = 1) = p$ for some fixed $p$, and let $\mathcal{P}$ be the set of distributions on $\mathcal{X} \times \mathcal{Y}$ with marginal $\Pr_Y$. Now the loss function takes three arguments, where $L(i, j, k)$ is the loss if $i$ is predicted, the true value of $Y$ is $j$, and $X = k$ is observed. Suppose that $L(i, j, k) = (k + 1)|i - j|$. That is, if the observation is 0, then, as before, the loss is just the difference between the predicted value and actual value; on the other hand, if the observation is 1, then the loss is twice the difference. Note that, with this loss function, it technically no longer makes sense to talk about ignoring the information, since we cannot even talk about the optimal rule with respect to $\Pr_Y$. However, as we shall see, the optimal action is still to predict the most likely value according to $\Pr_Y$.

A priori, there are four possible deterministic decision rules, which have the form "Predict $i$ if 0 is observed and $j$ if 1 is observed", which we abbreviate as $\delta_{ij}$, for $i, j \in \{0, 1\}$. It is easy to check that

$$E_{\Pr}[L_{\delta_{00}}] = \Pr(1, 0) + 2\Pr(1, 1) = \Pr_Y(1) + \Pr(1, 1)$$
$$E_{\Pr}[L_{\delta_{01}}] = \Pr(1, 0) + 2\Pr(0, 1)$$
$$E_{\Pr}[L_{\delta_{10}}] = \Pr(0, 0) + 2\Pr(1, 1)$$
$$E_{\Pr}[L_{\delta_{11}}] = \Pr(0, 0) + 2\Pr(0, 1) = \Pr_Y(0) + \Pr(0, 1).$$

It is not hard to show that randomization does not help in this case, and the minimax optimal decision rule is to predict 0 if $\Pr_Y(1) = p < 1/2$ and 1 if $p > 1/2$ (with any way of randomizing leading to the same loss if $p = 1/2$). Thus, the minimax-optimal decision rule still chooses the most likely prediction according to $\Pr_Y$, independent of the observation.

On the other hand, if either 0 or 1 is observed, the same arguments as before show that the minimax-optimal action with respect to the conditional probability is to randomize, predicting both 0 and 1 with probability $1/2$. So again, we have time inconsistency in the sense discussed in Section 2, and ignoring the information is the right thing to do.

But now consider what happens when the loss function is $L'(i, j, k) = (|k - j| + 1)|i - j|$. Thus, if the actual value and the observation are the same, then the loss function is the difference between the actual value and the prediction; however, if the actual value and the observation are different, then the loss is twice the difference. Again we have the same four decision rules as above, but now we have

$$E_{\Pr}[L'_{\delta_{00}}] = 2\Pr(1, 0) + \Pr(1, 1) = \Pr_Y(1) + \Pr(1, 0)$$
$$E_{\Pr}[L'_{\delta_{01}}] = 2\Pr(1, 0) + 2\Pr(0, 1)$$
$$E_{\Pr}[L_{\delta_{10}}] = \Pr(0, 0) + \Pr(1, 1)$$
$$E_{\Pr}[L_{\delta_{11}}] = \Pr(0, 0) + 2\Pr(0, 1) = \Pr_Y(0) + \Pr(0, 1).$$

Now the minimax-optimal rule is to predict 0 if $\Pr_Y(1) \leq 1/3$, to predict 1 if $\Pr_Y(1) \geq 2/3$, and to use the randomized decision rule $\frac{1}{3}\delta_{01} + \frac{2}{3}\delta_{10}$ (which has expected loss $2/3$) if $1/3 \leq \Pr_Y(1) \leq 2/3$. (In the case that $\Pr_Y(1) = 1/3$ or $\Pr_Y(1) = 2/3$, then the two recommended rules have the same payoff.)

On the other hand, if $i$ is observed ($i \in \{0, 1\}$), then the minimax optimal action is to predict $i$ with probability $1/3$ and $1 - i$ with probability $2/3$. That is, the optimal strategy corresponds to the decision rule $\frac{1}{3}\delta_{01} + \frac{2}{3}\delta_{10}$. Thus, in this case, there is no time inconsistency if $1/3 \leq \Pr_Y(1) \leq 2/3$. ∎

## 5 RELATED WORK

In this section we compare our work to recent and closely related work by Seidenfeld [2004] and Augustin [2003], as well as to various results indicating that information should never be ignored.

**Augustin's and Seidenfeld's work** Seidenfeld [2004] provides an analysis of minimax decision rules which is closely related to ours, but with markedly different conclusions. Suppose an agent has to predict the value of a random variable $Y$ after observing another random variable $X$. Seidenfeld observes, as we did in Section 2, that the minimax paradigm can be applied to this situation in two different ways:

1. In the *local* minimax strategy, the agent uses the minimax action relative to the set of distributions for $Y$ conditioned on the observed value of $X$.

2. In the *global* minimax strategy, the agent adopts the minimax decision *rule* (function from observations of $X$ to actions) relative to the set $\mathcal{P}$ of joint distributions.

Seidenfeld notes, as we did in Section 2, that the local minimax strategy is not equivalent to the global minimax strategy. (In his terminology, the extensive form of the decision problem is not equivalent to the normal form.) Moreover, he exhibits a rather counterintuitive property of local minimax. Suppose that, before observing $X$, the agent is offered the following proposition. For an additional small cost (loss), she will *not* be told the value of $X$ before she has to predict $Y$. An agent who uses the local minimax strategy would accept that proposition, because not observing $X$ leads to a smaller minimax prediction loss than observing $X$. Therefore, a local minimax agent would be willing to pay *not* to get information. This is the same phenomenon that we observed in Example 2.2.

Seidenfeld interprets his observations as evidence that the local minimax strategy is flawed, at least to some extent. He further views the discrepancy between local and global minimax as a problematic aspect of the minimax paradigm. In a closely related context, Augustin [2003] also observes the discrepancy between the global and the local minimax strategy, but, as he writes, "there are sound arguments for both".

In this paper, we express a third point of view: we regard both the strategy of ignoring information and the global minimax loss strategy as reasonable decision rules, preferrable to, for example, the local minimax loss strategy. However, we certainly do not claim that "global minimax loss" is the only reasonable strategy. For example, as we explained at the end of Section 3, in some situations minimax *regret* may be more appropriate. Also, as explained in Section 4, if $\mathcal{P}$ has a more complex structure than the one considered in Sections 2 and 3, then ignoring the information may no longer coincide with a global minimax strategy. It remains to be investigated whether, in such cases, there is a clear preference for either ignoring information or for global minimax.

**"Cost-free information should never be ignored"** As we observed in Section 3, a purely subjective Bayesian who is not "pragmatic" in our sense should always condition on *all* the available information: information should never be ignored. This result can be reconciled with our findings by noting that it depends on the agent representing her uncertainty with a *single* distribution. In the Bayesian case, the agent starts with a set of distributions $\mathcal{P}$, but this set is then transformed to a single distribution by adopting a subjective prior on $\mathcal{P}$. The expected value of information is then calculated using an expectation based on the agent's prior on $\mathcal{P}$. In contrast, in the "Bayesian" analysis of Section 3, for reasons explained at the beginning of Section 3, we computed the expectation relative to *all* probabilities in a set $\mathcal{P}$ that is meant to represent the agent's uncertainty. Consequently, our results differ from the subjective Bayesian analysis.

## 6 Discussion

We have shown that, in the minimax sense, sometimes it is better to ignore information, at least for a while, rather than updating. This strategy is essentially different from other popular probability updating mechanisms such as the non-Bayesian mechanism (local minimax) described in Section 3 and the Bayesian mechanism of Section 2. The only method we are aware of that leads to similar results is the following form of the maximum-entropy formalism: the agent first chooses the unique distribution $P^* \in \mathcal{P}$ that maximizes the Shannon entropy, and then predicts $Y$ based on the conditional distribution $P^*(Y = \cdot \mid X = x)$ [Cover and Thomas 1991]. Such an application of the Maximum Entropy Principle will ignore the value of $X$ if $\mathcal{P}$ contains *all* distributions with the given marginal $\Pr_Y$. However, as we indicated in Section 4, updating by ignoring can still be useful if $\mathcal{P}$ contains only a subset of the distributions with given $\Pr_Y$. Yet in such cases, it is well known that the maximum entropy $P^*$ may introduce counterintuitive dependencies between $X$ and $Y$ after all, as exemplified by the Judy Benjamin problems [Grove and Halpern 1997], thereby making the method different from merely ignoring $X$ after all. Our minimax-optimality results depend on the assumption that the set of possible prior distributions contains no information about the possible correlations between the variable of interest and the observed variable. In addition, they depend on the assumption that the payoff depends only on the actual value and the predicted value of the variable of interest.

One way of understanding the issues involved here is in terms of knowledge, as advocated by Halpern and Tuttle [1993], specifically, the knowledge of the agent and the knowledge of the "adversary" who is choosing the loss function. The knowledge of the agent is encoded by the set of possible prior distributions. The knowledge of the adversary is encoded in our assumptions on the loss function. If the adversary does not know the observation at the time that the loss function is determined, then the loss function cannot depend on the observation; if the adversary knows the observation, then it can. More generally, especially if negative losses (i.e., gains) are allowed, and the adversary can know the true distribution, then the adversary can choose whether to allow the agent to play at all, depending on the observation. In future work, we plan to consider the impact of allowing the adversary this extra degree of freedom.

**Acknowledgments**

We thank Teddy Seidenfeld and Bas van Fraassen for helpful discussions on the topic of the paper. Joseph Halpern was supported in part by NSF under grants CTC-0208535 and ITR-0325453, by ONR under grants N00014-00-1-03-41 and N00014-01-10-511, by the DoD Multidisciplinary University Research Initiative (MURI) program administered by the ONR under grant N00014-01-1-0795, and by AFOSR under grant F49620-02-1-0101.